\documentclass{article}

\usepackage[numbers,sort&compress]{natbib}
\usepackage[preprint]{neurips_2026}  %

\usepackage[utf8]{inputenc}
\usepackage[T1]{fontenc}
\usepackage{hyperref}
\usepackage{url}
\usepackage{booktabs}
\usepackage{amsfonts}
\usepackage{amssymb}
\usepackage{amsmath}
\usepackage{nicefrac}
\usepackage{microtype}
\usepackage[dvipsnames]{xcolor}
\usepackage{graphicx}
\usepackage{wrapfig}
\usepackage{soul}

\newcommand{\wkTitle}{Portable Causal Fairness Across \\
Synthetic Data Generator Families}

\newcommand{\wkVenue}{Trustworthy AI for Good (AI4GOOD)}

\newcommand{\wkAbstract}{%
When a statistical agency or regulator releases synthetic data in place of
sensitive records, it chooses the generator that produces the table, and can
shape that generator so unfair pathways are absent. DECAF made this concrete
on one non-private GAN: three fairness definitions become three sets of edge
cuts on the generator's causal graph. Whether the mechanism belongs to DECAF,
or to causal factorisation itself, was untested. We port all three definitions
to ten generators from four unrelated families (marginals-based, GAN, and
diffusion, each with differentially private variants, and the tabular foundation
model TabPFN), across three levels of formal privacy guarantee, over 2{,}640 matched-pair runs on Adult and COMPAS datasets.
The mechanism transfers everywhere. TabPFN is best overall: its release is the
most useful and loses the most disparity, though it starts with the most. Our new
causal diffusion backbone leaves the smallest gap under CF and has a private version. Applying the cut barely moves fidelity, costs a downstream classifier
about $0.08$ (CF) to $0.16$ (DemPar) AUC on average, and adding privacy
guarantees does not make the data less fair.%
}

\newcommand{\wkIntro}{%
Institutions holding sensitive records increasingly publish a synthetic
substitute rather than the records themselves: a hospital, a lending regulator
or a statistical agency fits a generative model and releases a table drawn from
it. The practice is growing: the 2020 US Census was released under a
differential-privacy guarantee \citep{abowd2022topdown}, and public challenges
run by NIST, the FDA and the US--UK privacy-technology programme have made
private tabular release a working technology rather than a proposal
\citep{nist2018challenge,vchamps2023,vos2024}. That
substitution is usually justified on privacy grounds, but it also lets the
publisher edit the process that generates the data. A disparity that exists in the
world because of a pathway running from a protected attribute, such as national
origin, to an outcome, such as whether a loan is approved, need not exist in
the released table.

The existing method closest to this is
DECAF~\citep{vanbreugel2021decaf}, which factors a generator along a causal
graph so that severing an edge at sampling time makes the generator
structurally unable to use the pathway it named. But DECAF is a single
non-private WGAN-GP~\citep{gulrajani2017wgangp}, whose released tables trail
modern differentially private marginals-based synthesizers and modern
diffusion models on the fidelity and utility benchmarks that decide what
institutions actually deploy (Sec.~\ref{sec:results}). Whether the fairness mechanism is confined to
that one architecture, or is a property of causal factorisation itself, is
therefore a practical question, since it decides whether an institution deploying synthetic data has the intervention available at all.

We ask two questions of that mechanism. Is edge surgery a property of DECAF,
or of the causal factorisation any such generator uses? And can it be attached
to a generator whose graph is not a supplied causal DAG but the undirected
marginal-dependency graph a differentially private synthesizer builds from
noisy counts? If the mechanism is confined to a
generator no institution would deploy, no released table will benefit from it.%
}

\newcommand{\wkDiscussion}{%

\textbf{Edge surgery belongs to causal factorisation, not to DECAF.} DECAF
conjectured that its mechanism would extend beyond its own GAN but tested no
other generator, and PreFair implements one such definition (CF) on a single
marginals synthesizer (MST). We show the definitions compile identically
across ten generators in four families, on both datasets and at every
privacy budget; an institution can now pick a synthesizer for its fidelity,
privacy or maturity and keep the fairness lever, instead of being pushed onto
a non-private GAN or a single marginals implementation to get it.

\textbf{TabPFN is the best overall for utility and fairness, and diffusion
leaves the smallest gap under CF.} Averaged over both datasets, TabPFN's
release is the most useful, and the cuts remove the most from its gap, which
starts largest. Diffusion leaves the smallest downstream
parity gap under CF ($0.099$), ahead of causal GAN ($0.128$), marginals
($0.152$) and TabPFN ($0.153$). Under DemPar it is level with TabPFN ($0.081$,
$0.080$), ahead of causal GAN ($0.089$) and marginals ($0.136$;
Table~\ref{tab:fairness}).
The diffusion backbone is new work: DECAF gave us one GAN and PreFair one marginals-based synthesizer, and
neither offered a diffusion route, through which CF turns out to leave the
least disparity.

\textbf{Which generator should a publisher use?} \emph{Marginals-based
synthesizers} remain the default when fidelity and privacy come first: among
the families that carry a privacy guarantee they are truest to the real table
on every fidelity measure, and they are private by construction.
\emph{Causal TabPFN} is the best overall choice when no formal guarantee is
needed. \emph{Causal diffusion} is
the choice when the gap after CF comes first. It has a private version and
holds fidelity close to the marginals tier. \emph{The
causal GAN family} trails on fidelity and utility, and DP-GAN fails
outright. On Adult, CF pays $6$ AUC points to remove $41\%$ of the baseline
gap on marginals, $16$ ($78\%$) on causal GAN, $18$ ($76\%$) on causal
diffusion, and $19$ ($60\%$) on TabPFN; the cost is smaller on COMPAS, where every family starts closer
to chance, and DemPar costs two to three times as much AUC as CF
on COMPAS. 

\textbf{Only the in-generator cut ships with the table.} On datasets where
the private structure encodes little to sever, a downstream mitigator can
reach a lower parity gap on the release (Table~\ref{tab:baselines}). But a
public table is used by consumers its publisher will never meet, and of the
four places to intervene, the in-generator cut is the only one that travels
with the release and cannot be dropped.

\textbf{Conclusion.} All four families share the same abstraction, a causal factorisation whose
edges can be severed, and the fairness mechanism belongs to that abstraction. An institution can now pick a synthesizer for
its privacy or data-quality needs, and keep a portable causal-fairness
intervention as part of that choice.%
}

\title{\wkTitle}
\workshoptitle{\wkVenue}

\author{%
  Steven Golob\thanks{Corresponding author.} \\
  University of Washington Tacoma \\
  \texttt{golobs@uw.edu} \\
  \And
  Sikha Pentyala \\
  University of Washington Tacoma \\
  \And
  Martine De Cock \\
  University of Washington Tacoma \\
}

\begin{document}

\maketitle

\begin{abstract}
\wkAbstract
\end{abstract}

\section{Introduction}

\wkIntro

\section{Background and related work}
\label{sec:related}

\begin{figure}[t]
  \centering
  \includegraphics[width=\textwidth]{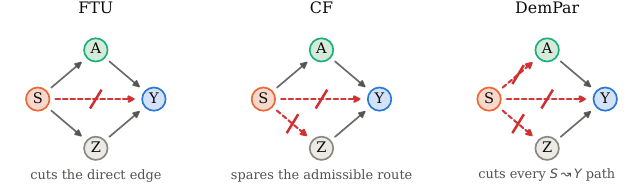}
  \caption{Three fairness definitions as three sets of severed edges, acting on
  the graph the data is \emph{generated} from, not on the model trained
  afterwards for a downstream task. $S$ is the protected attribute, $Y$ the outcome, $A$ an
  admissible attribute whose influence is deemed legitimate, and $Z$ an
  intermediate proxy. Dashed red edges are cut. DemPar removes every directed
  $S \rightsquigarrow Y$ path; CF removes only those that avoid $A$, so the
  route $S \rightarrow A \rightarrow Y$ survives.}
  \label{fig:pruning}
\end{figure}

\paragraph{DECAF: fairness as graph surgery.}
DECAF \citep{vanbreugel2021decaf} generates fair synthetic tables from a GAN
whose generator is \emph{factored} along a causal DAG (directed acyclic
graph): the joint distribution over columns is written as a product of
conditionals, $P(X) = \prod_i P(X_i \mid \mathrm{pa}(X_i))$, with
$\mathrm{pa}(X_i)$ the columns the graph draws as parents of column $X_i$.
This is the same decomposition a marginals-based synthesizer such as MST
\citep{mckenna2021mst} fits, differing chiefly in that MST's dependency
graph is undirected. DECAF gives each column its own sub-network reading
only $\mathrm{pa}(X_i)$, so a parent can be severed at generation time by
substituting a surrogate value drawn independently of the rest of the row.

Severing $X_j \rightarrow X_i$ re-fits nothing: the trained conditional is
instead \emph{evaluated} at a surrogate parent value resampled from $X_j$'s
own marginal, a \emph{do}-style intervention applied at sampling time. That is
why the cut is cheap, and also why it can cost fidelity: the sub-network is
asked about parent combinations it never saw in training.

Which edges to sever turns a fairness \emph{definition} into an
\emph{algorithm}, and DECAF supplies three (Figure~\ref{fig:pruning}).
\textbf{FTU} (fairness through unawareness) cuts only the direct edge $S
\rightarrow Y$; \textbf{DemPar}\footnote{DECAF calls this mechanism DP; we
write DemPar throughout and reserve DP for differential privacy, discussed
alongside it in every result below.} (demographic parity) cuts every directed
path $S \rightsquigarrow Y$; \textbf{CF} (conditional fairness) spares paths
that pass through an attribute the modeller has nominated \emph{admissible},
so disparity routed through a legitimate factor survives. DECAF's own
illustration is the clearest: $\textit{Education} \rightarrow \textit{Resume}
\rightarrow \textit{Job}$ is worth keeping because \textit{Resume} is
admissible, while $\textit{Race} \rightarrow \textit{Postcode} \rightarrow
\textit{Loan}$ is redlining and is not. DECAF's authors argue that the method should port to other
generators.

\paragraph{Differential privacy.}
Institutions releasing sensitive tables increasingly attach a differential
privacy (DP) guarantee \citep{dwork2006calibrating}: the 2020 US Census was
released under one \citep{abowd2022topdown}, and the synthesizers we study
here come out of the public challenges that made private
tabular release practical \citep{nist2018challenge,mckenna2021mst}.
DP caps the influence any single individual can have on the released table by
a budget $\varepsilon$: smaller $\varepsilon$ means stronger privacy and
noisier releases. $\varepsilon{=}1$ is the tight budget an institution might
publish under; $\varepsilon{=}1000$ is a permissive comparison point that
approximates no privacy at all. Every mention of ``private'' or ``DP-''
below refers to such a synthesizer.

\paragraph{Related approaches, and the gap.}
Only PreFair \citep{pujol2023prefair} attaches an in-generator fairness cut to
a private synthesizer, adapting CF to MST alone by excluding offending
marginals from candidacy before the private structure search runs. Everything
else combining privacy with fairness in generation acts outside the graph:
enforcing independence on a learned embedding
\citep{hyrup2025flip,sarmin2025pfwgan}, prompting an LLM backbone
\citep{nagesh2025faircausesyn}, learning a private structure without a
fairness objective \citep{jia2025pradagan}, or mitigating downstream
\citep{angelozzi2026where}. Causal-DAG diffusion targets fidelity only
\citep{zhang2025causaldifftab,jacob2026tabscm}, and private tabular diffusion
without a causal graph exposes no per-column edge to cut
\citep{zhu2024dptldm}. We benchmark against \citet{angelozzi2026where} in
\S\ref{sec:downstream}, the one prior system on the same units (parity gap on
a held-out real population); the embedding- and LLM-based generators are
outside our scope, which is whether in-generator edge surgery transfers across
families rather than a head-to-head between fair-SDG algorithms.

So one question is open: is DECAF's edge surgery a property of the
complete \emph{DECAF} architecture, or of \emph{causal factorisation}?
DECAF conjectures the latter, calling the method ``simple and extendable to
other generative methods'', but never tests it. The stakes are practical: if
the mechanism is confined to DECAF, a publisher who wants a fair release must
accept a non-private GAN that, on our own measurements
(Figure~\ref{fig:results}), no institution would deploy; if it is a property
of the factorisation, the publisher keeps the fairness lever while picking a
synthesizer for its privacy and quality, spanning the marginals-based
synthesizers that dominate private tabular data
\citep{mckenna2021mst,zhang2014privbayes,zhang2021privsyn} and diffusion,
which leads several tabular-quality benchmarks \citep{kotelnikov2023tabddpm}.

\section{Porting the mechanisms}
\label{sec:method}

We implement FTU, CF and DemPar once, against a single
interface,\footnote{Code: \url{https://github.com/steveng9/CausalFairnessInSDG}.}
and port that interface to ten generators from four unrelated families. The
fairness definitions do not change. What changes is the graph the cut acts on,
and whether that graph is directed or undirected. Each definition reduces to
a graph-separation constraint on the release's dependency structure (no
unadmissible $S \rightsquigarrow Y$ path); DECAF's directed-edge severance and
our undirected-pair exclusion are two implementations of the same constraint,
so the DAG in DECAF's presentation is a design choice.

\textbf{Marginals-based} synthesizers (MST \citep{mckenna2021mst},
PrivBayes \citep{zhang2014privbayes} and PrivSyn \citep{zhang2021privsyn}) form a graphical model
whose edges are the pairs of columns whose noisy joint counts the synthesizer
chooses to measure under the privacy budget; the release is drawn from a
distribution reproducing those counts. Their graph is undirected, so we
enforce the cut on \emph{paths} rather than on ordered
parent\,$\rightarrow$\,child edges. Whether adding a candidate pair would
create a forbidden $S$-to-$Y$ path depends on what the model already
contains, so we check the constraint incrementally at each selection step. A pair the cut excludes is never measured, and the check reads only the
role declaration and the structure already chosen, never the data (following
PreFair \citep{pujol2023prefair} for CF on MST, extended here to PrivBayes,
PrivSyn, and to DemPar and FTU).

\textbf{Causal GANs} (DECAF, plus variants we build on a CTGAN
representation \citep{xu2019ctgan} and on DP-SGD \citep{abadi2016dpsgd})
take DECAF's surrogate substitution unchanged, on the DAG. For
\textbf{causal diffusion} we keep the same factorisation and replace each
sub-network with a small denoiser trained by supervised noise prediction. The
cut is then the same code path, because what exposes an edge is that $X_i$ has
its own network reading only $\mathrm{pa}(X_i)$, true whether that network
was trained adversarially, as in the GAN families, or by denoising.
\textbf{Causal TabPFN} keeps the factorisation too, and takes each conditional
from the tabular foundation model TabPFN \citep{hollmann2025tabpfn}: the real
training rows are its context, with $\mathrm{pa}(X_i)$ as features, and we
sample from the probabilities it returns. Nothing is trained, and a severed
parent takes its value from a randomly permuted row, as in the other causal
backbones.

\paragraph{Experimental setup.}
Two datasets, Adult \citep{becker1996adult} and COMPAS
\citep{angwin2016machinebias}; ten generators; four mechanisms
(\textsc{none}/FTU/CF/DemPar); three protected/admissible role splits per
dataset (Appendix~\ref{app:roles}); five seeds; $2{,}640$ completed runs. Six of
the ten generators carry a differential-privacy guarantee
\citep{dwork2006calibrating} and are marked $\dagger$ throughout: the three
marginals-based synthesizers, which are private by construction, and the three
backbones to which we attach DP-SGD (DP-GAN, DP-CTGAN, DP-Diffusion). Those
six run at $\varepsilon \in \{1,10,1000\}$ with $\delta = 10^{-9}$; DECAF,
$+$CTGAN, $+$Diffusion and $+$TabPFN have no privacy mechanism and run once. For the
causal GAN, diffusion and TabPFN families we specify each dataset's DAG by hand from
domain knowledge, as is standard in this literature (following
\citet{kusner2017counterfactual} and \citet{zhang2017causalframework}), and
hold it fixed across generators so that any effect is attributable to the
mechanism. The marginals-based generators choose their own undirected graph
under the privacy budget, and we enforce the cut on that graph.

\paragraph{The fairness metric.}
Every fairness number in this paper is a \textbf{demographic-parity gap}.
Train a classifier on the released synthetic table, use it to predict the
outcome for a held-out sample of \emph{real} people, and compare how often it
says yes to each protected group:
\[
  \mathrm{gap} \;=\; \bigl|\;\Pr(\hat{Y}{=}1 \mid S{=}1)\;-\;\Pr(\hat{Y}{=}1 \mid S{=}0)\;\bigr| ,
\]
reported as the worst case over the protected attributes of the split. For
example, if a classifier trained on a synthetic Adult table predicts
above-\$50k income for $50\%$ of the men and $17\%$ of the women in the real
held-out sample, the gap is $0.33$; a mechanism that brings it to $0.05$ has
removed $85\%$ of the gap. The gap asks only whether the two groups receive the positive prediction equally often, not
whether either prediction is correct. We use it because it is the disparity DemPar is defined to
remove, and because it is measured on real people, so a generator cannot score
well by inventing a fairer population than the one it was given. We also
record conditional parity and positive- and negative-rate balance for every
run. CF and DemPar lower all four in every family on both datasets, and we
report the parity gap because it is the one the three mechanisms are defined
against.

\paragraph{Fidelity, utility, and matched pairs.}
For \emph{fidelity} we report total-variation distance (TVD) between
synthetic and real marginals ($0$ = agree exactly, $1$ = share no mass),
averaged over single columns ($1$-way) and column pairs ($2$-way), plus the
mean absolute difference in Cram\'er's $V$ across column pairs (which asks
whether the release preserves the \emph{strength} of each pairwise
association). For \emph{utility} we train a classifier on the synthetic table
and score it on the real held-out $30\%$, reporting AUC (floor $0.5$) and
macro-F1 (floor $0.41$) against a train-on-real reference. Because seed noise
rivals the fairness effects themselves, every fairness number below is a
\textbf{matched-pairs} difference: each mechanism run is differenced against
the run identical in dataset, generator, $\varepsilon$, role split \emph{and}
seed, but with no mechanism.

\section{Results}
\label{sec:results}

\begin{figure}[t]
  \centering
  \includegraphics[width=\textwidth]{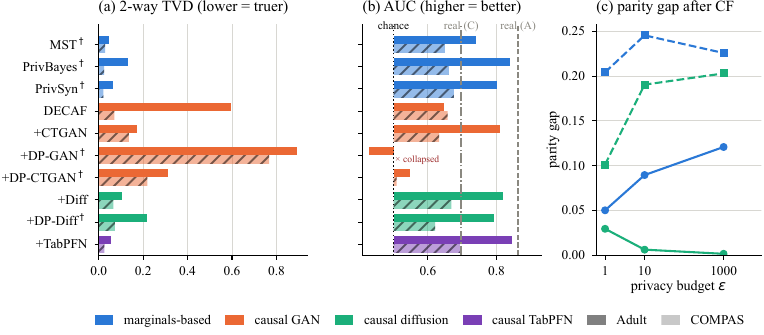}
  \caption{\textbf{(a,\,b)} every generator with no fairness mechanism,
  averaged over role splits and seeds; (a) $2$-way TVD to the real table
  (lower is better), (b) downstream AUC on real held-out people (higher is
  better), with chance ($0.5$) and train-on-real references marked; a cell
  where the release collapsed to a single outcome class carries no AUC and
  is marked as such. \textbf{(c)} the parity gap remaining after CF at each
  privacy budget, for the two families with a matched cell at every
  $\varepsilon$ (marginals-based and causal diffusion; the causal GAN family
  has DP-GAN and DP-CTGAN collapses that leave some cells empty, and TabPFN
  has no private version). Each line
  is one family--dataset pairing (Adult solid, COMPAS dashed). $\dagger$
  marks a generator carrying a differential-privacy guarantee.}
  \label{fig:results}
\end{figure}

\subsection{Fidelity and utility}

Fidelity separates the ten generators cleanly on $2$-way TVD
(Figure~\ref{fig:results}a). The marginals-based synthesizers do well, as
expected since they optimise almost exactly this metric: MST reaches $0.047$
on Adult and $0.028$ on COMPAS. TabPFN matches them without a privacy
guarantee ($0.057$ and $0.026$). Among the trained causal backbones, our
diffusion backbone is next: it reaches $0.103$ on Adult against DECAF's own $0.597$, and with DP-SGD
attached, DP-diffusion at $0.218$ still leads DP-CTGAN at $0.313$ and DP-GAN
at $0.895$.

Utility separates the same way (Figure~\ref{fig:results}b). Most generators
clear the AUC floor of $0.5$ comfortably on both datasets; on Adult,
TabPFN reaches $0.846$, PrivBayes $0.839$, causal diffusion $0.819$, DP-diffusion $0.791$
at $\varepsilon{=}10$, MST $0.742$. \textbf{Only DP-GAN collapses}: AUC
$0.427$ on Adult (below chance) and a constant predictor on COMPAS, an
artefact of DP-SGD on a WGAN-GP discriminator that we diagnose in
\S\ref{sec:privacy}. A collapsed generator has no disparity for a fairness
mechanism to remove; since every fairness number below is a matched-pair
difference, this only affects DP-GAN's own row.

\subsection{Fairness findings}
\label{sec:fairness}

\begin{table}[t]
  \caption{Matched-pairs effect of each mechanism on the worst-case
  demographic-parity gap, pooled within generator family, restricted to trials
  where the no-mechanism release lets a downstream random forest reach
  $\mathrm{AUC}>0.6$ (so that ``fairness'' is not a trivial artefact of
  collapsed learning). \emph{gap} and \emph{AUC} are what remains after the
  mechanism, both measured on real held-out people; the parenthesised value
  beside each is the mean \emph{paired} change against the identical run with
  no mechanism (negative $=$ the gap shrank), with a $95\%$ CI on the gap.
  Each family header gives its no-mechanism gap and AUC on
  Adult\,/\,COMPAS. $\dagger$ marks a generator carrying a differential-privacy
  guarantee; those run at $\varepsilon{=}10$.}
  \label{tab:fairness}
  \centering
  \footnotesize
  \setlength{\tabcolsep}{4pt}
  \begin{tabular}{l cc c cc}
    \toprule
    & \multicolumn{2}{c}{\textbf{Adult}} & & \multicolumn{2}{c}{\textbf{COMPAS}} \\
    \cmidrule(lr){2-3} \cmidrule(lr){5-6}
    Mech. & gap ($\Delta$) & AUC ($\Delta$) & & gap ($\Delta$) & AUC ($\Delta$) \\
    \midrule
    \multicolumn{6}{l}{\textit{Marginals-based} (MST$^\dagger$, PrivBayes$^\dagger$, PrivSyn$^\dagger$); none: gap 0.210 / 0.234, AUC 0.794 / 0.662} \\
    FTU    & 0.192 ($-0.019 \pm 0.022$) & 0.795 ($+0.002$) & & 0.192 ($-0.042 \pm 0.026$) & 0.666 ($+0.004$) \\
    CF     & 0.125 ($-0.086 \pm 0.024$) & 0.737 ($-0.057$) & & 0.178 ($-0.056 \pm 0.033$) & 0.644 ($-0.018$) \\
    DemPar & 0.103 ($-0.108 \pm 0.027$) & 0.640 ($-0.154$) & & 0.168 ($-0.066 \pm 0.034$) & 0.626 ($-0.036$) \\
    \addlinespace[2pt]
    \multicolumn{6}{l}{\textit{Causal GAN} (DECAF, $+$CTGAN, $+$DP-GAN$^\dagger$, $+$DP-CTGAN$^\dagger$); none: gap 0.164 / 0.254, AUC 0.770 / 0.653} \\
    FTU    & 0.144 ($-0.020 \pm 0.034$) & 0.762 ($-0.008$) & & 0.216 ($-0.039 \pm 0.032$) & 0.649 ($-0.004$) \\
    CF     & 0.036 ($-0.128 \pm 0.033$) & 0.612 ($-0.158$) & & 0.219 ($-0.035 \pm 0.038$) & 0.620 ($-0.033$) \\
    DemPar & 0.018 ($-0.147 \pm 0.032$) & 0.535 ($-0.235$) & & 0.160 ($-0.094 \pm 0.048$) & 0.554 ($-0.099$) \\
    \addlinespace[2pt]
    \multicolumn{6}{l}{\textit{Causal diffusion} ($+$Diffusion, $+$DP-Diffusion$^\dagger$); none: gap 0.159 / 0.218, AUC 0.805 / 0.647} \\
    FTU    & 0.159 ($-0.001 \pm 0.009$) & 0.805 ($+0.000$) & & 0.156 ($-0.063 \pm 0.036$) & 0.633 ($-0.014$) \\
    CF     & 0.039 ($-0.121 \pm 0.010$) & 0.622 ($-0.183$) & & 0.159 ($-0.060 \pm 0.036$) & 0.601 ($-0.045$) \\
    DemPar & 0.025 ($-0.135 \pm 0.011$) & 0.493 ($-0.312$) & & 0.137 ($-0.081 \pm 0.043$) & 0.515 ($-0.132$) \\
    \addlinespace[2pt]
    \multicolumn{6}{l}{\textit{Causal TabPFN} ($+$TabPFN); none: gap 0.200 / 0.338, AUC 0.846 / 0.693} \\
    FTU    & 0.200 ($-0.001 \pm 0.008$) & 0.842 ($-0.003$) & & 0.226 ($-0.111 \pm 0.030$) & 0.676 ($-0.017$) \\
    CF     & 0.081 ($-0.119 \pm 0.011$) & 0.655 ($-0.191$) & & 0.225 ($-0.113 \pm 0.035$) & 0.633 ($-0.060$) \\
    DemPar & 0.056 ($-0.144 \pm 0.011$) & 0.489 ($-0.357$) & & 0.104 ($-0.234 \pm 0.024$) & 0.536 ($-0.157$) \\
    \bottomrule
  \end{tabular}
\end{table}

\textbf{The definitions transfer.} Cutting edges reduces the downstream
parity gap in every family, on both datasets, at every privacy level
(Table~\ref{tab:fairness}); the predicted ordering (DemPar $>$ CF $>$ FTU by
amount removed) holds in seven of eight family$\times$dataset cells, a sanity
check on the definitions themselves. Nothing about the three definitions
requires a GAN, and nothing requires a supplied DAG.

\textbf{Averaged across datasets, the cuts remove the most disparity from
TabPFN, and diffusion is left with the smallest gap under CF.} TabPFN starts
from the largest gap and loses $0.116$ of it under CF and $0.189$ under
DemPar (diffusion: $0.090$ and $0.108$). The mean parity gap remaining on real held-out
data after CF is $0.099$ for diffusion, $0.128$ for causal GAN, $0.152$
for marginals and $0.153$ for TabPFN; under DemPar, $0.081$, $0.089$, $0.136$
and $0.080$
(Table~\ref{tab:fairness}). On Adult, diffusion cuts an initial gap of
$0.159$ to $0.039$ under CF and $0.025$ under DemPar: a release whose
downstream classifier predicts the positive outcome at almost equal rates
for the two protected groups. CF removes $76\%$ of the baseline gap on
diffusion (Adult), $78\%$ on causal GAN, $60\%$ on TabPFN and $41\%$ on marginals; on Adult the share
removed is larger for the three families whose generators follow a causal DAG.

\textbf{The two graph types differ in cost, not in what they cut.} How many
$S \rightsquigarrow Y$ edges each graph contains varies by dataset
(Table~\ref{tab:edges}); more consistently, per edge severed a DAG cut costs
about $2.5\times$ as much downstream utility as a marginals cut, because a
marginals cut excludes a pair before training while a DAG cut asks a trained
sub-network to extrapolate (Appendix~\ref{app:per_edge_cost}).

\begin{table}[t]
  \caption{Mean number of edges the mechanism must sever, per run, per
  generator family. Causal GAN/diffusion/TabPFN counts are deterministic (DAG edges
  DECAF's surrogate substitution would replace under each dataset's role
  split); marginals counts are the minimum edge cut on the baseline learned
  graph needed to break every forbidden $S$-to-$Y$ path.}
  \label{tab:edges}
  \centering
  \footnotesize
  \setlength{\tabcolsep}{6pt}
  \begin{tabular}{l ccc c ccc}
    \toprule
    & \multicolumn{3}{c}{\textbf{Adult}} & & \multicolumn{3}{c}{\textbf{COMPAS}} \\
    \cmidrule(lr){2-4} \cmidrule(lr){6-8}
    Family & FTU & CF & DemPar & & FTU & CF & DemPar \\
    \midrule
    GAN / diffusion / TabPFN (fixed DAG) & 0.75 & 4.00 & 8.75 & & 1.33 & 1.67 & 3.33 \\
    Marginals-based (learned graph)   & 0.58 & 2.81 & 4.24 & & 0.96 & 2.27 & 3.72 \\
    \bottomrule
  \end{tabular}
\end{table}

\textbf{The cut hardly moves fidelity.} Pooled across all ten generators
and both datasets, CF shifts the released table's $2$-way TVD to the real
distribution by $0.001$ and DemPar by $0.003$, against a
no-fairness-mechanism TVD that ranges from $0.04$ to $0.83$ across
generators; the release is essentially as faithful after the cut as
before. The utility cost is modest under CF (about $0.08$ downstream AUC on
average) and larger under DemPar (about $0.16$; Table~\ref{tab:fairness}).

\subsection{What TabPFN adds}
\label{sec:tabpfn}

On \emph{utility}, TabPFN's release is the
most useful of the ten: AUC $0.846$ on Adult and $0.693$ on COMPAS, against
$0.863$ and $0.697$ for training on real data. On \emph{fairness}, it starts from the largest
COMPAS gap ($0.338$), and DemPar removes the most there ($0.234$).
On \emph{privacy}, it has no differential-privacy guarantee, since its context
is the real table. Under DemPar its downstream AUC falls to $0.489$ on Adult
and $0.536$ on COMPAS, so a small gap there comes with a classifier near
chance, as it does on diffusion. In a larger study of 11 datasets, where CF's cut
set is chosen by d-separation, we measured how much of the gap lost accuracy
alone would close, by scrambling the release's labels until the classifier
loses as much AUC as the cut costs. On TabPFN, this noise closes $98\%$ of the
gap that DemPar closes and $82\%$ of CF's (diffusion: $96\%$ and $75\%$).

\subsection{How this compares to intervening downstream}
\label{sec:downstream}

\begin{table}[t]
\centering
\footnotesize
\setlength{\tabcolsep}{7pt}
\renewcommand{\arraystretch}{0.93}
\caption{\textbf{Where to intervene, on identical data.} Absolute parity gap
on MST synthetic training data at differential-privacy budget $\varepsilon$,
real held-out test set, XGBoost, 5 seeds, all inside the benchmark harness
of \citet{angelozzi2026where}. The eight upper mitigators act on the
released table or the classifier trained from it; our three
(\textbf{in-gen (ours)}) act inside the generator. The harness reports
accuracy but not AUC or macro-F1, and every method scores within a point of
the majority baseline ($0.76$ Adult, $0.53$ COMPAS), so accuracy does not
discriminate methods and we omit it. $\ddagger$ marks a row where the
classifier collapsed further, below majority.}
\label{tab:baselines}
\begin{tabular}{@{}ll@{\hspace{30pt}}cccc@{}}
\toprule
 & & \multicolumn{2}{c}{Adult} & \multicolumn{2}{c}{COMPAS} \\
\cmidrule(lr){3-4}\cmidrule(lr){5-6}
stage & intervention & $\varepsilon{=}1$ & $\varepsilon{=}10$ & $\varepsilon{=}1$ & $\varepsilon{=}10$ \\
\midrule
 & no mitigation                     & 0.068 & 0.073 & 0.109 & 0.190 \\
\addlinespace[2pt]
pre  & Reweighing                    & 0.087 & 0.085 & 0.132 & 0.093 \\
pre  & Disp.\ impact rem.            & 0.078 & 0.105 & 0.119 & 0.205 \\
pre  & Learned fair repr.$^\ddagger$ & 0.018 & 0.031 & 0.037 & 0.028 \\
\addlinespace[2pt]
in   & Exp.\ gradient                & 0.047 & 0.058 & 0.115 & 0.053 \\
in   & Grid search                   & 0.068 & 0.073 & 0.109 & 0.190 \\
\addlinespace[2pt]
post & Reject option                 & 0.045 & 0.046 & 0.045 & 0.046 \\
post & Equalised odds                & \textbf{0.009} & 0.015 & \textbf{0.024} & \textbf{0.023} \\
post & Calib.\ eq.\ odds             & 0.049 & 0.040 & 0.485 & 0.415 \\
\addlinespace[2pt]
\textbf{in-gen (ours)} & FTU         & 0.073 & 0.062 & 0.087 & 0.155 \\
\textbf{in-gen (ours)} & CF          & 0.015 & 0.029 & 0.111 & 0.155 \\
\textbf{in-gen (ours)} & DemPar      & 0.010 & \textbf{0.006} & 0.067 & 0.148 \\
\bottomrule
\end{tabular}
\end{table}

Everything above measures the cut against itself. To place it against the
alternatives an institution already has, we run the benchmark of
\citet{angelozzi2026where}, which asks where in a synthesis pipeline a
fairness intervention should go and supplies eight standard answers, all
acting on the released synthetic table or the classifier trained from it
(pre-processing the table, constraining the trained classifier, or
post-processing its predictions). We add a fourth stage (in-generator)
by running MST-FTU, MST-CF and MST-DemPar as extra rows in the same
benchmark, otherwise unmodified: its preprocessed data, splits, MST
synthesizer, XGBoost classifier, aif360 metrics, and its default role split
(matching PreFair; Appendix~\ref{app:roles}).

On Adult, our in-generator DemPar reaches a parity gap of $0.006$ at
$\varepsilon{=}10$, close to the strongest downstream mitigator
(equalised-odds post-processing at $0.015$) and lower than every pre- and
in-processing entry. On COMPAS, equalised-odds post-processing wins ($0.023$
vs.\ our $0.148$); our cut still improves fairness over no mitigation.

A downstream mitigator can be dropped, altered or replaced by whoever trains
on the release, while an in-generator cut ships with the table itself
(\S\ref{sec:discussion}).

\subsection{Privacy and fairness are nearly independent}
\label{sec:privacy}

A tighter DP budget might be expected to blunt fairness cuts, since DP
suppression falls hardest on the smallest groups
\citep{bagdasaryan2019disparate}, but we do not observe that. Pooled across
the six DP generators, the parity gap after CF sits at $0.09$--$0.13$ at
every $\varepsilon \in \{1, 10, 1000\}$ we tested
(Figure~\ref{fig:results}c), and after DemPar at $0.08$--$0.12$; the
mechanism removes whatever disparity the budget produced.

\textbf{One private generator fails outright.} DECAF's post-processing
argument invites swapping in a DP-SGD \citep{abadi2016dpsgd} discriminator;
doing so collapses the synthetic outcome to a single class in $168/360$ runs
($47\%$). Two substitutions each remove
the failure (CTGAN representation, $1/360$; diffusion in place of the
WGAN-GP discriminator, $0/360$), locating the cause in DP-SGD on a
weight-clipped discriminator, not in private causal generation as such.

\section{Discussion}
\label{sec:discussion}

\begin{wrapfigure}{r}{0.44\textwidth}
  \vspace{-\baselineskip}
  \centering
  \includegraphics[width=0.44\textwidth]{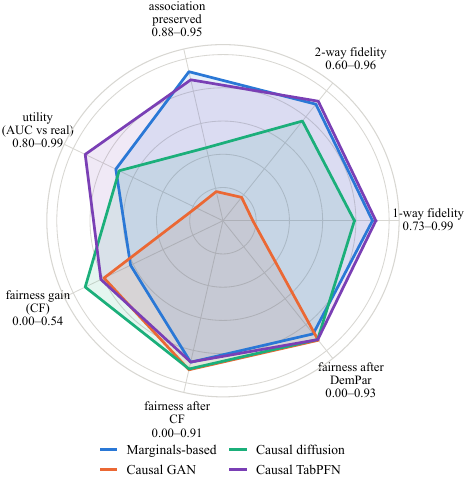}
  \caption{The four generator families on seven axes, averaged over both
  datasets, role splits and seeds ($\varepsilon{=}10$ for private
  generators); higher is better. TVD, Cram\'er's $V$ difference and parity gap
  are plotted as one minus their value. Fairness axes start at $0$, the others
  at the weakest family. \emph{Fairness gain} is the share of the baseline gap
  that CF removes. Collapsed runs count in the two \emph{fairness after} axes,
  which raises the causal GAN there.}
  \label{fig:radar}
\end{wrapfigure}

\wkDiscussion

\section*{Acknowledgments}
This material is based upon work supported by the National Science Foundation
under Grant No.~2523406. Steven Golob is supported by an NSF CSGrad4US
Fellowship. Sikha Pentyala is affiliated with the eScience Institute.

\bibliographystyle{plainnat}
\bibliography{refs}

\appendix

\section{Role splits}
\label{app:roles}

\begin{table}[h]
  \caption{The three role splits per dataset used throughout the paper.
  \emph{Protected} is $S$, \emph{admissible} is $A$ (the attributes CF is
  allowed to route disparity through), and the outcome $Y$ is
  \texttt{income} on Adult and \texttt{two\_year\_recid} on COMPAS. The
  first split of each dataset follows PreFair's Table~1; the others widen
  and narrow the admissible set, since a narrow admissible set should make
  CF behave more like DemPar.}
  \label{tab:roles}
  \centering
  \footnotesize
  \setlength{\tabcolsep}{4pt}
  \begin{tabular}{@{}l p{0.28\textwidth} p{0.50\textwidth}@{}}
    \toprule
    Split & Protected $S$ & Admissible $A$ \\
    \midrule
    \multicolumn{3}{l}{\textit{Adult}} \\
    PreFair       & \texttt{sex}, \texttt{race}, \texttt{native-country} & \texttt{workclass}, \texttt{education}, \texttt{occupation}, \texttt{hours-per-week}, \texttt{capital-gain}, \texttt{capital-loss} \\
    Broad         & \texttt{sex} & \texttt{workclass}, \texttt{education}, \texttt{education-num}, \texttt{occupation}, \texttt{hours-per-week} \\
    Narrow        & \texttt{sex}, \texttt{race} & \texttt{education-num}, \texttt{hours-per-week} \\
    \addlinespace[2pt]
    \multicolumn{3}{l}{\textit{COMPAS}} \\
    PreFair       & \texttt{sex}, \texttt{race} & \texttt{c\_charge\_degree}, \texttt{priors\_count} \\
    Broad         & \texttt{race} & \texttt{c\_charge\_degree}, \texttt{priors\_count}, \texttt{juv\_fel\_count}, \texttt{juv\_misd\_count} \\
    Narrow        & \texttt{sex}, \texttt{race}, \texttt{age\_cat} & \texttt{c\_charge\_degree} \\
    \bottomrule
  \end{tabular}
\end{table}

\section{Per-edge cost of the cut}
\label{app:per_edge_cost}

Table~\ref{tab:edges} in \S\ref{sec:fairness} shows the two graph types
sever different numbers of edges to enforce the same fairness definition.
Here we normalise both fairness gain and utility cost by that count, and
find that the two graph types deliver essentially the same fairness per
severed edge but differ substantially in the utility they pay per edge.

\begin{table}[h]
  \caption{Fairness gain and utility cost under CF, normalised by the
  number of edges CF severs (Table~\ref{tab:edges}). Values are matched-pair
  means across seeds, role splits and privacy budgets, restricted to trials
  with baseline $\mathrm{AUC}>0.6$.}
  \label{tab:per_edge}
  \centering
  \footnotesize
  \begin{tabular}{l cc cc}
    \toprule
    & \multicolumn{2}{c}{gap reduction / edge} & \multicolumn{2}{c}{$\Delta$AUC / edge} \\
    \cmidrule(lr){2-3} \cmidrule(lr){4-5}
    Graph & Adult & COMPAS & Adult & COMPAS \\
    \midrule
    DAG (GAN, diffusion)   & 0.031 & 0.028 & 0.043 & 0.024 \\
    Marginals (undirected) & 0.031 & 0.025 & 0.020 & 0.008 \\
    \bottomrule
  \end{tabular}
\end{table}

Under CF, the gap-per-edge is essentially the same across graph types
($\approx 0.03$), so per severed edge the two mechanisms are equally
effective at removing disparity. The utility cost per edge, in contrast, is
about $2.5\times$ larger on the DAG (mean $0.033$ AUC per edge across
datasets) than on the marginals graph (mean $0.014$). Under DemPar the
direction on utility cost agrees on three of four cells but is diluted on
Adult, where the DAG severs $\approx 9$ edges and the per-edge denominator
grows.

A mechanism-level reading follows from \S\ref{sec:method}. Severing an edge
on the DAG evaluates a trained sub-network on parent combinations it never
saw at training time, so the fidelity hit propagates into every column
downstream of that sub-network. Severing an edge on the marginals graph
merely excludes a candidate pair from the graphical model's structure search
before training begins, so the model is fit on the remaining constraints
without ever being asked to extrapolate. This is why the DAG cut is cheap but not free (\S\ref{sec:related}).

On Adult, the
private marginals graph carries the direct protected-to-outcome edge in none
of the $15$ matched pairs across seeds and generators, so the FTU cut is a
literal no-op in every one: the definition is satisfied by the graph the
private structure search selected under the budget, before any fairness
constraint is imposed. Private marginals graphs are not always fair, but in this case the choice of
graph removed the direct edge on its own.

\end{document}